\documentclass{article}
\usepackage[final]{nips_2018}
\usepackage{amsmath,amssymb,amsfonts}
\usepackage{algorithmic}
\usepackage{graphicx}
\usepackage{textcomp}
\usepackage{xcolor}
\PassOptionsToPackage{numbers}{natbib}
\usepackage{hyperref}   
\usepackage[cal=boondoxo,calscaled=1]{mathalfa}
\usepackage[rmdefault=ntxmi,sfdefault=iwona,scaled=1.0]{isomath}
\DeclareSymbolFont{letters}{OML}{ntxmi}{m}{it}

\def\BibTeX{{\rm B\kern-.05em{\sc i\kern-.025em b}\kern-.08em
    T\kern-.1667em\lower.7ex\hbox{E}\kern-.125emX}}
\begin{document}

\title{Benchmarking Deep Sequential Models on \\ Volatility Predictions for Financial Time Series
}

 \author{
   Qiang Zhang$^\dag$,  Rui Luo$^\dag$, Yaodong Yang, Yuanyuan Liu \vspace{.1cm}\\
   American International Group Inc.\thanks{Presented at NIPS 2018 Workshop on \emph{Challenges and Opportunities for AI in Financial Services}. $^\dag$ First two authors contributed equally. Correspondence to Yuanyuan Liu: $<$yuanyuan.liu@aig.com$>$. }
}
\maketitle

\begin{abstract}
Volatility is a quantity of measurement for the price movements of stocks or options which indicates the uncertainty within financial markets. As an indicator of the level of risk or the degree of variation,  volatility is important to  analyse the financial market, and it is taken into consideration in various decision-making processes in financial activities. 
On the other hand,  recent advancement in  deep learning techniques has shown strong capabilities in modelling sequential data, such as speech and natural language. 
In this paper, we empirically study the applicability of the latest deep structures with respect to the volatility modelling problem, through which we aim to provide an empirical guidance for the theoretical analysis of the marriage between deep learning techniques and financial applications in the future. 
We examine both the traditional approaches and the deep sequential models on the task of volatility prediction, including the most recent variants of convolutional and recurrent networks, such as the dilated architecture. 
Accordingly, experiments with real-world stock price datasets are performed on a set of 1314 daily stock series for 2018 days of transaction. 
The evaluation and comparison are based on the negative log likelihood (NLL) of real-world stock price time series. 
The result shows that the dilated neural models, including dilated CNN and Dilated RNN, produce most accurate estimation and prediction, outperforming various widely-used deterministic models in the GARCH family and several recently proposed stochastic models. 
In addition, the high flexibility and rich expressive power are validated in this study.

\end{abstract}


\section{Introduction}
Volatility is a quantity of measurement for the price movements of stocks or options which indicates the uncertainty within financial markets. 
As an indicator of  the level of risk or the degree of variation, the volatility is of great importance in the analysis of financial markets, and it is taken into consideration before investment decisions are made and portfolios are optimised \citep{hull2006options}. Moreover, it is essentially a key variable in the pricing of derivatives. 
Therefore, the estimation of volatility is of fundamental importance in many of the branches of financial studies, such as investment, risk management, security valuation and monetary policy making \citep{poon2003forecasting}.

As a quantitative measurement, volatility is quantified typically via  standard deviation of the change of price in a fixed time interval, for example in a day, month or year. 
The level of risk is roughly proportional with the magnitude of  volatility. 
We have to establish reasonable and reliable volatility models to recognise the patterns within the financial time series. 
Among the various challenges in designing volatility models, one of the primary challenges is determining the existence of hidden random processes and to characterise the underlying dependences or interactions among different variables within a certain time span. 
One of the conventional approaches is utilising the domain knowledge to choose the characteristic features of volatility models and impose assumptions and constraints on the models based on the observation of historical data and specification of the task. 
Notable examples include autoregressive conditional hetero-scedasticity (ARCH) model \citep{engle1982autoregressive} and its extension, generalised ARCH (GARCH)  \citep{bollerslev1986generalized}, which take advantage of the autoregression operations to extract the properties of time-varying volatility within many time series. 
The family of GARCH model has been expanding since the inception of the two aforementioned models. As an alternative to the GARCH model family, the class of stochastic volatility (SV) models consider the variance of the time series follow a certain hidden random process \citep{hull1987pricing}.
Heston \citep{heston1993closed} introduced a continuous-time model with the stochastic volatility following an Ornstein-Uhlenbeck process which helps to derive a closed-form solution for options pricing. 
Avellaneda et al. \citep{avellaneda1996managing} used a volatility band to model hetero-scedasticity and rectify the solution under the worst-case volatility scenarios.
Since the time discretisation of continuous-time dynamics sometimes results in a deviation from the original trajectory of system, those continuous-time models are  rarely applied in forecasting because time steps in forecasting tasks are always discrete. 
Hence, for forecasting practically, the canonical model  \citep{jacquier2002bayesian,kim1998stochastic} which is presented in a discrete-time manner for regularly spaced series, such as daily prices of stocks or options is substantially interesting.
Although theoretically sound, those methods require rather strong assumptions which might be involving detailed insights of the time series to investigate and are hence hard to analyse without a comprehensive inspection.

The progress in deep learning has led to great success in many of areas of research, for instance image recognition \citep{DBLP:conf/nips/KrizhevskySH12,DBLP:conf/cvpr/HeZRS16}, speech recognition \citep{DBLP:conf/nips/ChorowskiBSCB15}, machine translation \citep{DBLP:journals/corr/BahdanauCB14,DBLP:conf/emnlp/LuongPM15}, and generating models for images or text \citep{DBLP:journals/corr/Graves13,DBLP:journals/corr/KingmaW13,DBLP:conf/icml/RezendeMW14,DBLP:conf/icml/GregorDGRW15}. 
Deep learning technique is essentially the creative exploitation of deep neural networks with various complex structures and different nonlinear activation functions as a highly flexible nonlinear models to fit the target function. 
Particularly, for sequential learning, Chung et al. \citep{DBLP:conf/nips/ChungKDGCB15} and Fraccaro et al. \citep{DBLP:conf/nips/FraccaroSPW16} have proposed stochastic recurrent structures for sequential modelling where the uncertainty is to some extent maintained by the stochastic layers within the neural networks.

As per the recent advancement to the stochastic volatility models, some efforts have been made to integrate the successes of deep learning techniques from other areas into  volatility modelling, such as the stochastic volatility model with stochastic recurrent neural networks and variational inference by Luo et al. \citep{DBLP:conf/aaai/LuoZXW18}. 
The central property is that it takes a fully data-driven approach to determine the configurations with the least exogenous input as possible: a neural network re-formulation of stochastic volatility is proposed by leveraging stochastic models and recurrent neural networks (RNNs).




In this paper, we empirically study the applicability of the latest deep structures with respect to the volatility modelling problem. 
We benchmark traditional approaches and deep sequential models in terms of volatility prediction. 
Additionally, experiments with real-world stock price datasets are performed. The result shows that the dilated neural models, including dilated CNN and dilated RNN, produce more accurate estimation and prediction, outperforming various widely-used deterministic models in the GARCH family and several recently proposed stochastic models on average negative log-likelihood (NLL). Consequently, the high flexibility and rich expressive power are validated.


\section{Related Work}

It has been decades since the inception of volatility models \citep{engle2001good}. During these decades of research, volatility modelling is proved to be an effective technique for accurate forecasts within financial markets \citep{andersen1998answering}. 
As a result, many models have been proposed. A notable example for volatility modelling is the Nobel prize-winning model -- the autoregressive conditional hetero-scedasticity (ARCH) model  \citep{engle1982autoregressive}: it is capable of providing accurate identification of the characteristics of time-varying volatility within various types of time series. 
Apart from the ARCH model, a large body of diverse work has emerged based on stochastic processes for volatility modelling  \citep{bollerslev1994arch}. The most famous one is   Bollerslev's generalised ARCH model \citep{bollerslev1986generalized} which generalises autoregressive conditional hetero-scedasticity (GARCH) model so that it is analogous to the extension from the autoregressive (AR) model to the autoregressive moving average (ARMA) model by introducing the past conditional variances in the current conditional variance estimation. 
Engle and Kroner  \citep{engle1995multivariate} demonstrate theoretical results on the formulation and estimation of multivariate GARCH model within simultaneous equation systems. The extension to multivariate model allows the covariance to present and depend on the historical information, which is particularly useful in multivariate financial models. 
One of the alternatives to the conditionally deterministic GARCH model family is the class of stochastic volatility (SV) models, as first appeared in the theoretical finance literature on option pricing  \citep{hull1987pricing}. The SV models regard variance to be following some certain hidden stochastic process so that the current volatility is no longer a deterministic function even though the historical information is given. 
An example is Heston's model  \citep{heston1993closed} which characterises the variance process as a Cox-Ingersoll-Ross (CIR) process driven by a latent Wiener process. This enables a closed-form solution analytically for the ease of option pricing or strategy design. 
Although theoretically sound, these models require strong assumptions in order to function properly which involves assuming complex probability distributions and nonlinear driving dynamics. 
Moreover, for practically implementing these models, one might have to impose less prior knowledge and rectify a solution under the worst-case volatility case  \citep{avellaneda1996managing}. 
Nevertheless, models, such as ARCH and its descendants/variants, have been confirmed to provide accurate prediction \citep{andersen1998answering}, consequently, becoming indispensable tools in asset or option pricing and risk evaluation/management. 
Notably, several models have been recently proposed for practical forecasting tasks. 
Kastner et al.  \citep{DBLP:journals/csda/KastnerF14} exploited the framework \emph{stochvol} based on MCMC where the ancillarity-sufficiency interweaving strategy (ASIS) is applied for boosting MCMC estimation of stochastic volatility. 
Wu et al.  \citep{DBLP:conf/nips/WuHG14} proposed the \emph{GP-Vol} as a non-parametric model which uses Gaussian processes to identify the dynamics and jointly learns the volatility process and hidden states through online inference algorithm.
 Luo et al. \citep{DBLP:conf/aaai/LuoZXW18} introduced a neural network formulation to the stochastic volatility modelling as a general framework  where stochastic autoregressive layers, bidirectional recurrent neural networks and variational inference techniques are leveraged to provide  a more systematic and volatility-specific formulation on stochastic volatility modelling for the ease of learning the complex non-linear dynamics and its internal uncertainty. 

Although these models  give us a practically viable route towards stochastic volatility forecasting, they need a considerable amount of training samples to guarantee the models' accuracy, which involves very expensive sampling routine at each time step. Another drawback is that they are incapable of handling the forecasting task for multivariate time series. 

On the other hand, deep learning  \citep{DBLP:journals/nature/LeCunBH15,DBLP:journals/nn/Schmidhuber15} that exploits nonlinear structures, known as the deep neural networks, powers various applications. 
It overcomes the pattern recognition challenges, such as image recognition \citep{DBLP:conf/nips/KrizhevskySH12,DBLP:conf/cvpr/HeZRS16}, speech recognition \citep{DBLP:conf/nips/ChorowskiBSCB15}, machine translation \citep{DBLP:journals/corr/BahdanauCB14,DBLP:conf/emnlp/LuongPM15}, and generating models for images or text \citep{DBLP:journals/corr/Graves13,DBLP:journals/corr/KingmaW13,DBLP:conf/icml/RezendeMW14,DBLP:conf/icml/GregorDGRW15} to name a few.

As one of the time-dependent network models, RNN has been extensively integrated in advanced structures, such as long short-term memory (LSTM)  \citep{DBLP:journals/neco/HochreiterS97}, bidirectional RNN (BRNN)  \citep{DBLP:journals/tsp/SchusterP97}, gated recurrent unit (GRU)  \citep{DBLP:conf/emnlp/ChoMGBBSB14} and attention mechanism  \citep{DBLP:conf/icml/XuBKCCSZB15}.
Recent results show that RNNs excel in sequence modelling and generation of various applications  \citep{van2016pixel,DBLP:conf/emnlp/ChoMGBBSB14,DBLP:conf/icml/XuBKCCSZB15}. 
However, despite its capability as non-linear universal approximator, one of the drawbacks of neural networks is its deterministic nature. Adding latent variables and their processes into neural networks could easily make the posteriori computationally intractable. Recent research has demonstrated that efficient inference can be found using variational inference when hidden continuous variables are embedded into the neural networks structure  \citep{DBLP:journals/corr/KingmaW13,DBLP:conf/icml/RezendeMW14}. 
Some researchers  have started to explore the use of variational inference to make RNNs stochastic, for example 
Chung et al.  \citep{DBLP:conf/nips/ChungKDGCB15} defined a sequential framework with complex interacting dynamics of coupling observable and latent variables, whereas Fraccaro et al.  \citep{DBLP:conf/nips/FraccaroSPW16} used heterogeneous backward propagating layers in inference network according to its Markovian properties.

Generally, RNNs are considered while modelling sequential data, such as financial time series. 
Deep recurrent networks are often applied in complicated sequential pattern recognition tasks. On one hand, they provide us with a viable route to complex patterns. On the other hand, they have  training issues, including the vanishing or exploding gradient, low parallelism, etc.; these issues get worsen as the depth of networks increases. 
The difficulties within training deep recurrent networks have been known for  a long time \citep{pascanu2013difficulty}, and lot of efforts have been devoted for neural networks to enhance the training stability or robustness of deep recurrent neural networks. A lot of the latest models are attempting to address some of these long-standing problems

The latest architectures of CNN and RNN have achieved the state-of-the-art result within many areas of research. The temporal convolutional networks (TCN) \citep{DBLP:journals/corr/abs-1803-01271} is a sequential convolutional architecture that guarantees the network out to be of the same length with the input, and information leakage from future is interdicted to the past. 
The dilated recurrent neural networks (DilatedRNN) \citep{chang2017dilated} have been designed to use  multi-resolution dilated recurrent skip connections; dilated recurrent layers are stacked for the extraction of complex data dependencies, and the dilation increases exponentially across layers. 
The independent recurrent neural networks (IndRNN) \citep{DBLP:journals/corr/abs-1803-04831} takes the raw data as well as its previous hidden state and output as the current input to  solve the gradient vanishing problem. 
The quasi-recurrent neural network (QRNN) \citep{DBLP:journals/corr/BradburyMXS16} essentially alternates between the convolutional layers and recurrent pooling layers. 
It aims at enhancing the parallelism by decoupling the dependencies between consecutive time steps. The skip recurrent neural networks (SkipRNN) \citep{DBLP:journals/corr/abs-1708-06834} extends existing recurrent neural networks by learning to skip the state updates with no explicit identification of whether a sample is informative to the current task. 
The recurrent highway network (RHN) \citep{zilly2017recurrent} adapts the so-called highway layers \citep{DBLP:journals/corr/SrivastavaGS15} to a recurrent setting. The highway layers exploits adaptive training for very deep feedforward neural networks which enable larger step-to-step transition, mitigating the training difficulty. 
The hierarchical multi-scale recurrent neural network (HM-RNN) \citep{DBLP:journals/corr/ChungAB16} aims at modelling complex temporal dependencies by determining hierarchical patterns within time series. A key component of HM-RNN is the parameterised boundary detector (PBD), which generates a binary value in each hidden layer; HM-RNN relies on PBD to determine the time for finalising a segment for optimising the overall objective. 
The fast-slow recurrent neural network (FS-RNN) \citep{mujika2017fast} inherits advantages from both HM-RNN and the RHN. FS-RNN processes information at different time resolution and learns complex temporal transition functions.


\section{Task Description}
The task of sequential volatility prediction is to conduct 1-step-ahead prediction of volatility value given the history values. Formally, given a series of sequential data $X = (x_1, x_2, \dots, x_T)$, the sequential model is to predict the value of $x_{T+1}$.

These models are evaluated according to the NLL between the 1-step-ahead prediction and the label value. Thus the loss function can be defined as follows: 
\begin{equation}
L = - \log \frac{1}{\sqrt{2\pi \sigma^2}}\exp(-\frac{x^2}{\sqrt{2 \sigma^2}}),
\end{equation}
where $\sigma$ is the 1-step-ahead model prediction.

\section{Sequential Neural Models}
\label{baseline}


This section details the baselines to model volatility time series. Recent advancement in convolutional and recurrent neural networks have been included here.

\textbf{Temporal Convolutional Network (TCN)} \citep{DBLP:journals/corr/abs-1803-01271} is a convolutional architecture for sequential modelling. TCN is reported to achieve higher evaluation metrics and parallelism than recurrent networks in several sequential modelling tasks. TCN mainly consists of two designing principles: the network out has the same length with the input, and information leakage from future is interdicted to the past. 

1D-fully-convolutional network (FCN) and casual convolution are deployed to achieve the above-mentioned two points, respectively. FCN adds zero padding to the hidden layers in order to keep subsequent layers of the same length as pervious layers. Casual convolutions guarantee that the computation of the current time step can only access information from current input and previous hidden states. 
One major disadvantage of casual convolutions is that we need to construct very deep networks and very quite large filter banks in order to model complicated sequential data dependencies. 

Mathematically,  the 1D-dilated convolutional operation $F$ on element $s$ of a sequence can be defined as follows:
\begin{equation}
F(s) = (\mathbf{x} \ast _{d}f)(s) = \sum_{i=0}^{k-1}f(i)\cdot \mathbf{x}_{s-d\cdot i},
\end{equation}
where $d$ is the dilation factor, $k$ is the filter size and $s-d\cdot i$ represents data from the past.

TCN enjoys several advantages of sequence modelling. First, TCN has flexible receptive field size and can change it in different approaches. For example, it can increase the size of filters and the dilation factor in each hidden layer; the number of hidden layers can also efficiently increase receptive field size and strengthen the model's memory capability. Second, gradients can back propagate along different temporal directions, and thus ease the problem of exploding and vanishing gradients. Besides, TCN has high parallelism as convolutions can be conducted in parallel and avoid waiting for predecessors to complete computation.

\textbf{Dilated Recurrent Neural Network (DilatedRNN)} \citep{chang2017dilated} is characterised by multi-resolution dilated recurrent skip connections. In order to extract complex data dependencies, dilated recurrent layers can be stacked and the dilation increases exponentially across layers. DilatedRNN is thus more suitable for modelling very long-term dependencies. The dilated skip connection can be described as follows:
\begin{equation}
c_t^l = f(x_t^l, c_{t-s^l}^l),
\end{equation}
where $c$ denotes the cell state, $l$ is the layer and $t$ denotes the time step.

Dilated recurrent skip connections allow information to spread along fewer edges, which reduces the number of model parameters and eases the problem of exploding and vanishing gradients. Stacked DilatedRNN makes different layers focus on different resolution along the time axis and increases capability of modelling long-term dependencies. 

As compared to dilated convolutional neural networks (e.g., TCN) with the same number of layers and dilation rate in each layer, the DilatedRNN has the same number of recurrent edges per node which indicates that these two architectures have the same model complexity. 
However, the DilatedRNN is able to run beyond two cycles, especially with GRU and LSTM. Hence, DilatedRNN is reported to have stronger memory capability than TCN. \citep{chang2017dilated}
 
\textbf{Independently Recurrent Neural Network (IndRNN)} \citep{DBLP:journals/corr/abs-1803-04831} is designed to solve the gradient vanishing and exploding problem. Different from the traditional RNN that is observed as a multilayer perceptron over time, the IndRNN only receives information from inputs and its own previous time step. The neurons at the same layer are disconnected and thus independent from each other. This suggests that one neuron merely focusses on one type of spacial pattern. IndRNN can be formally described as follows:
\begin{equation}
h_{n,t} = \sigma(\mathbf{w}_{n}\mathbf{x}_t + u_n  h_{n,t-1} + b_n),
\end{equation}
where $\mathbf{w}_{n}$ and $u_n$ are the $n$-th row of the input weight matrix and recurrent weight matrix, respectively.

Correlation among different neurons can be reflected in the  connection across layers. By stacking the basic IndRNN layers, a deep IndRNN network could be constructed. Top hidden layers can aggregate different spatial patterns from lower hidden layers which could increase the capability of exploring complicated data dependencies. 
Compared to  stacked GRU with $\mathrm{Sigmoid}$ as an activation function, the stacked IndRNN can use non-saturated activation functions, such as $\mathrm{ReLU}$, and thus avoid the concern of decaying gradient caused by sigmoid. \citep{DBLP:journals/corr/abs-1803-04831} Batch normalisation can also be directly applied before or after the activation functions.

\textbf{Quasi-Recurrent Neural Network (QRNN)} \citep{DBLP:journals/corr/BradburyMXS16} alternates between convolutional layers and recurrent pooling functions. It is targeted at improving  low parallelism due to the dependencies of computation at current time step on that of previous time steps. Stacked QRNN is reported to have better prediction accuracy than stacked LSTM with the same hidden structure. 

Each layer of QRNN consists of two types of operations: convolution and recurrent pooling, which are quite analogous to convolutional layers and pooling layers in typical CNN architectures. The convolutional operation is conducted along the time axis with filter banks. Masked convolution \citep{van2016pixel} is adopted to prevent information leakage from future. Filter banks are specifically not allowed to gain future information for any computation of the given time steps. Additional separate filter banks are used in recurrent pooling functions to create element-wise gates, such as forget gates and output gates, at each time step. Dynamic average pooling \citep{balduzzi2016strongly} is applied to mix multiple states across time steps. Suppose the pooling function requires a forgetting gate $\mathbf{f}_t$ and an output gate $\mathbf{o}_t$ at each time step, the full computational procedure can be described as follows:
\begin{eqnarray}
\begin{split}
\mathbf{Z} &= \tanh(\mathbf{W}_z \ast \mathbf{X})\\
\mathbf{F} &= \sigma(\mathbf{W}_f \ast \mathbf{X})\\
\mathbf{O} &= \sigma(\mathbf{W}_o \ast \mathbf{X}),\\
\end{split}
\end{eqnarray}
where $\mathbf{W}_z$, $\mathbf{W}_f$ and $\mathbf{W}_o$ are the convolutional filters and $\ast$ denotes a masked convolution along the time direction.

The most notable advantage of QRNN is that many existing extensions to convolutional networks and recurrent networks can be directly applied to QRNN. Regularisation, such as variational inference-based dropout \citep{gal2016theoretically} and zoneout, \citep{DBLP:journals/corr/KruegerMKPBKGBL16} could be a robust extension inspired from LSTM.


\textbf{Skip Recurrent Neural Network (SkipRNN)} \citep{DBLP:journals/corr/abs-1708-06834} enriches existing recurrent networks by learning to skip state update without explicit information about which samples are useless for the current task. 
SkipRNN is optimised to shorten effective size of computational graph and conduct fewer updates for long-term dependency modelling. A binary state update gate, $u_t \in \{0,1\}$, is proposed to decide whether the RNN will update the states or merely copy the previous states. 
At each time step $t$, the sequential model learns to emit the probability of updating the states, i.e. $\hat{u}_{t+1} \in [0, 1]$. The whole procedure can be demonstrated as follows:
\begin{eqnarray}
\begin{split}
u_t &= f_{binarize}(\hat{u}_t)\\
s_t &= u_t \cdot S(s_{t-1}, x_t) + (1-u_t) \cdot s_{t-1}\\
\Delta \hat{u}_t &= \sigma(W_p s_t + b+p)\\
\hat{u}_{t+1}& = u_t \cdot \Delta \hat{u}_t + (1-u_t) \cdot (\hat{u}_t + \min (\Delta \hat{u}_t, 1-\hat{u}_t)),\\
\end{split}
\end{eqnarray}
where $W_p$ is the weight vector, $b_p$ is the bias, and $ f_{binarize}: [0,1] \rightarrow \{0, 1\}$.

Advantages of learning to skip state updates lie in several aspects. First of all, fewer updating steps back propagate the gradient further. Copying previous states increases the network memory and its capability to model long-term sequential dependencies. Besides, fewer state updates indicate less computational operation and faster convergence speed which makes model training much easier than other RNN variants. Furthermore, the learning-to-skip technique is orthogonal to recent advances in recurrent neural networks, and thus can be jointly used with other techniques, such as normalisation \citep{DBLP:journals/corr/CooijmansBLC16,DBLP:journals/corr/BaKH16} and regularisation \citep{DBLP:journals/corr/ZarembaSV14,DBLP:journals/corr/KruegerMKPBKGBL16}.   

One limitation of SkipRNN is the trade-off between the model performance and its total number of processed samples. One might need to sacrifice the model accuracy points due to some application circumstances \citep{DBLP:journals/corr/CooijmansBLC16}.

\textbf{Recurrent Highway Network (RHN)} \citep{zilly2017recurrent} incorporates highway layers \citep{DBLP:journals/corr/SrivastavaGS15} to recurrent transition. They use adaptive computation to train very deep feedforward neural networks. Incorporation of highway layers enables larger step-to-step transition which can ease the difficulty of training RNN models with deep transition functions. The highway layer computation can be demonstrated as follows:
\begin{equation}
\mathbf{y} =  \mathbf{h} \cdot \mathbf{t} + \mathbf{x} \cdot \mathbf{c},
\end{equation}
where $\mathbf{h}=H(\mathbf{x}, \mathbf{W}_H)$, $\mathbf{t}=H(\mathbf{x}, \mathbf{W}_T)$, $\mathbf{c}=H(\mathbf{x}, \mathbf{W}_C)$ are non-linear transform results with weights $\mathbf{W}_H$, $\mathbf{W}_T$ and $\mathbf{W}_C$ respectively, and ``$\cdot$'' represents element-wise multiplication between vectors.

Based on the Gersgorin circle theorem, LSTM variants have direct mechanism to efficiently regulate their Jacobian eigenvalues across time steps, which makes them more powerful than other RNN variants in modelling complicated sequential data. It has further been theoretically proved that LSTM is essentially a simplified version of RHN, but RHN possesses stronger memory capability to learn complex sequence processing.

\textbf{Hierarchical Multi-scale Recurrent Neural Network (HM-RNN)} \citep{DBLP:journals/corr/ChungAB16} aims to model complex temporal dependencies by discovering hierarchical structure of time series data. 
A major element in the HM-RNN model is the parameterised boundary detector (PBD) that generates a binary value in each hidden layer. 
HM-RNN relies on the PBD to decide when to end a segment in order to optimise the overall objective. 
With the state of the boundary, each hidden layer selects one of the three operations at each time step: COPY, UPDATE and FLUSH.

For a HM-RNN model with $L$ layers, the update procedure at time $t$ in the layer $l$ can be described as follows:
\begin{equation}
\mathbf{h}_t^l, \mathbf{c}_t^l, z_t^l = f^l_{HM-RNN}(\mathbf{c}_{t-1}^l, \mathbf{h}_{t-1}^l, \mathbf{h}_t^{l-1}, \mathbf{h}_{t-1}^{l+1}, z_{t}^{l-1}),
\end{equation}
where $z$ is the boundary state, $\mathbf{h}$ and $\mathbf{c}$ denote hidden and cell states, respectively.

The COPY operation simply duplicates hidden states at previous time steps and keeps them un-changed until it receives a summarised input from the lower layers. When the boundary states in the lower layer is detected and the boundary in the current layer is not found, the UPDATE operation will be conducted to change the summarised representations of this current layer. If a boundary is found, then the FLUSH operation will be conducted. The following two sub-operations are then executed: the EJECT sub-operation passes the current states to the upper layer and the RESET sub-operation re-initialises hidden states in the current layer before reading a new segment.


\textbf{Fast-Slow Recurrent Neural Network (FS-RNN)} \citep{mujika2017fast} inherits advantages from both HM-RNN and RHN. FS-RNN processes information at different time resolution and learns complex transition functions from previous time steps to the next time step. 

The FS-RNN architecture consists to $k$ sequentially connected Fast RNN cells $F_1, F_2, \dots, F_k$ on the lower hierarchical hidden layer and one Slow RNN cell $S$ on the higher hierarchical layer. The lower layer is thus called the Fast layer and the higher layer is called the Slow layer. The basic architecture with arbitrary RNN cells can be described through following equations:
\begin{eqnarray}
\begin{split}
h_t^{F_1} &= f^{F_1}(h_{t-1}^{F_k}, x_t)\\
h_t^S& = f^S(h_{t-1}^S, h_t^{F_1} )\\
h_t^{F_2} &= f^{F_2}(h_{t}^{F_1}, h_t^S)\\
h_t^{F_i} &= f^{F_i}(h_{t}^{F_i-1}) \qquad  \rm{for} \quad 3 \leq \it {i} \leq k \\
\end{split}
\end{eqnarray}

Essentially, the Fast cells have limited capability to model long-term sequence dependencies. With help of the Slow cells, shorter gradient propagation paths are added to the model and distant dependency between sequential inputs can be  tracked more efficiently. Hence, the FS-RNN inherits advantages from multi-scale RNN and deep transition RNN. 

Note that any RNN cells, such as LSTM and GRU, could be the basic computation unit of the FS-RNN architecture.

\section{Experiments}

\subsection{Dataset and Pre-processing}
The models are evaluated on real-world stock price time series data to validate performance. The raw dataset contains 1555 univariate time series of the daily closing stock price. It was recorded for 2018 working days. Missing value or insufficient observation of the stock price series introduces noise and can be highly influential to model performance. As these factors are not relevant to the ultimate purpose of volatility modelling, we filter the stock price time series that have missing value or insufficient observation. After filtering the raw dataset, we get 1314 stock daily price time series as the cleaned dataset. Then, we transform the original stock price $s_t$ into log-returns $x_t=log(s_t/s_{t-1})$ and normalise the transformed series. We split the normalised dataset along the time axis into two subsets: the first 1800 (~80\%) time steps of each time series data make up the training subset and the remaining 213 (~20\%) time steps of each series make up the testing subset.

\subsection{Model Implementation}
We evaluate and compare all convolutional and recurrent models described in Section \ref{baseline}. TCN is designed to consist of two hidden layers of  the kernel size 5. The dilation factor in each hidden layer is set to 2. For normalisation, we normalise weights of every convolutional filter \citep{salimans2016weight}. DilatedRNN sets the starting dilation to 1 and the dilation factor to 2. RHN consists of the carry gate that is coupled to the transformed gate by setting $C(\cdot)=\mathrm{1}_n-T(\cdot)$. All recurrent models adopt GRU as their basic architecture, except that RHN which adopts LSTM.

\begin{table*}[t!]
\label{tab:comparison}
\caption{Performance comparison of the proposed model with baselines with regard to negative log-likelihood (NLL) on the test subset.}
\begin{center}
\resizebox{1.01\textwidth}{!}{
\begin{tabular}{cccccccccccc}
\hline
Stock  & TCN             & DilatedRNN    & IndRNN & QRNN  & SkipRNN & HM-RNN & FS-RNN & RHN & ARCH & GARCH & EGARCH\\
\hline
1        &\textbf{1.335} & 1.645             & 1.859      & 1.593    & 1.462       & 1.683        &1.808       &1.864   &2.413&  1.794&1.402  \\
2        &\textbf{1.415} & 1.812             & 1.784      & 1.985    & 1.646      & 1.604        & 1.927       &1.896   & 1.865& 1.624 &1.501  \\
3        &1.669             & 1.641              & 1.845      & 1.722    & \textbf{1.612}      & 2.117        &1.750        &1.820 &2.521 &2.512  &1.641   \\
4        &2.117             & \textbf{1.599}              & 1.926     & 1.666    & 1.689      & 1.727        &2.665       & 1.823 & 1.875&1.867  &1.706   \\
5        &\textbf{1.405} & 1.587             & 1.829      & 1.715    & 1.758      &1.694         &1.722        &1.848    &2.169& 2.194  &1.775 \\
6        &\textbf{1.405} & 1.625             & 1.781      & 1.494    & 1.955      & 1.764 	       &1.628       &1.999   &1.622 & 1.569  &1.546   \\
7        &1.759             & 1.675             & 1.835      & 1.841    & 1.813      & 1.718		 &\textbf{1.498}         &1.833  &2.898  &2.472 &2.301 \\
8        &1.683             & 1.605             & 1.840      & \textbf{1.536}    & 1.667      &1.769   	 &1.916        &1.830  & 2.651&  2.571  &2.198     \\
9        &1.923             & 1.725             & 1.721      & 1.664    & 1.697      & \textbf{1.650} 	&1.6711       &1.841 & 3.431& 3.243   &3.293    \\
10      &2.324             & \textbf{1.475}             & 1.852      & 1.773    & 1.673      & 1.743  	&1.563        &1.796 &2.363  & 2.547 &3.654     \\
\hline
AVG   &\textbf{1.901} & 1.903          & 2.033    & 1.908    &1.925      &2.121  		&2.451       & 2.044 & 2.432  & 2.084 & 2.003\\
\hline
\end{tabular}}
\end{center}
\end{table*}

A multi-layer perceptron projects the last hidden layer to one-dimensional output. The default activation function in hidden layers and output layers is the Rectified Linear Unit ($\mathrm{ReLU}$). All weight matrices are initialised by standard normal distribution. Adam \citep{DBLP:journals/corr/KingmaW13} is selected to optimise model parameters. We also apply dropout \citep{srivastava2014dropout} gradient clipping \citep{pascanu2013difficulty} and learning rate annealing to help train deep models. Model hyper-parameters are determined by five-fold cross validation. The default batch size is set to 64. Unless specified otherwise, all models are implemented in PyTorch \citep{paszke2017automatic} and run on a single NVIDIA TITAN XP GPU.


\section{Results and Discussion}

The experimental results and performance are listed in Table~\ref{tab:comparison}. Evaluation is based on negative log-likelihood of 1-step-ahead prediction on each series in the testing subset. The baselines are compared on ten randomly selected stock price series. The average NLL of all the 1314 series are also reported in Table~\ref{tab:comparison}. 

Fundamentally, dilated architectures, including TCN and DilatedRNN, achieve the least NLL loss for the sequential prediction on all the stock test data. This indicates that dilated neural networks have higher flexibility and stronger expressive capability for volatility modelling and forecasting. Specifically, the dilated convolutional neural network (i.e., TCN) obtains slightly better performance than the dilated recurrent neural network (i.e., DilatedRNN). It can also be concluded that convolutional models are usually more suitable for volatility forecasting than recurrent models, as TCN out-performs all of the RNN that are studied. Furthermore, SkipRNN and QRNN achieve marginally worsen than others but quite similar performance to each other on the whole testing subset, while FS-RNN achieves the worst performance in volatility modelling and forecasting.       

These baselines behave differently when modelling the randomly selected 10 stock time series data. TCN achieves four ``first'' places in forecasting Stock 1, Stock 2, Stock 5 and Stock 6 while DilatedRNN beats other models for two stock series of data. In order to compare the best four neural models more efficiently, we select Stock 1, Stock 3, Stock 4 and Stock 8, where these four models have achieved the least loss, respectively. Figure \ref{fig:merged_figure} in the Appendix illuminates the performance of baselines on the selected four stock series.

\section{Conclusion}
Deep learning has been making advances in modelling sequential data such as natural language and speech. This paper focusses on benching recent deep learning models on volatility forecasting. Experimental results demonstrate that dilated neural networks including dilated CNN and dilated RNN have a better capability and outperform other models. We, therefore, model volatility alongside the dilated architectures.

\clearpage
\section*{Disclaimer}
This is not AIG work product and the paper expresses the views of the authors only in their individual capacity and not those of American International Group, Inc. or its affiliate entities.

{
\bibliographystyle{aaai-named}
\bibliography{volatility}
}
\clearpage
\section*{Appendix}

\begin{figure}[h!]

\centering
\includegraphics[width=.82\textwidth]{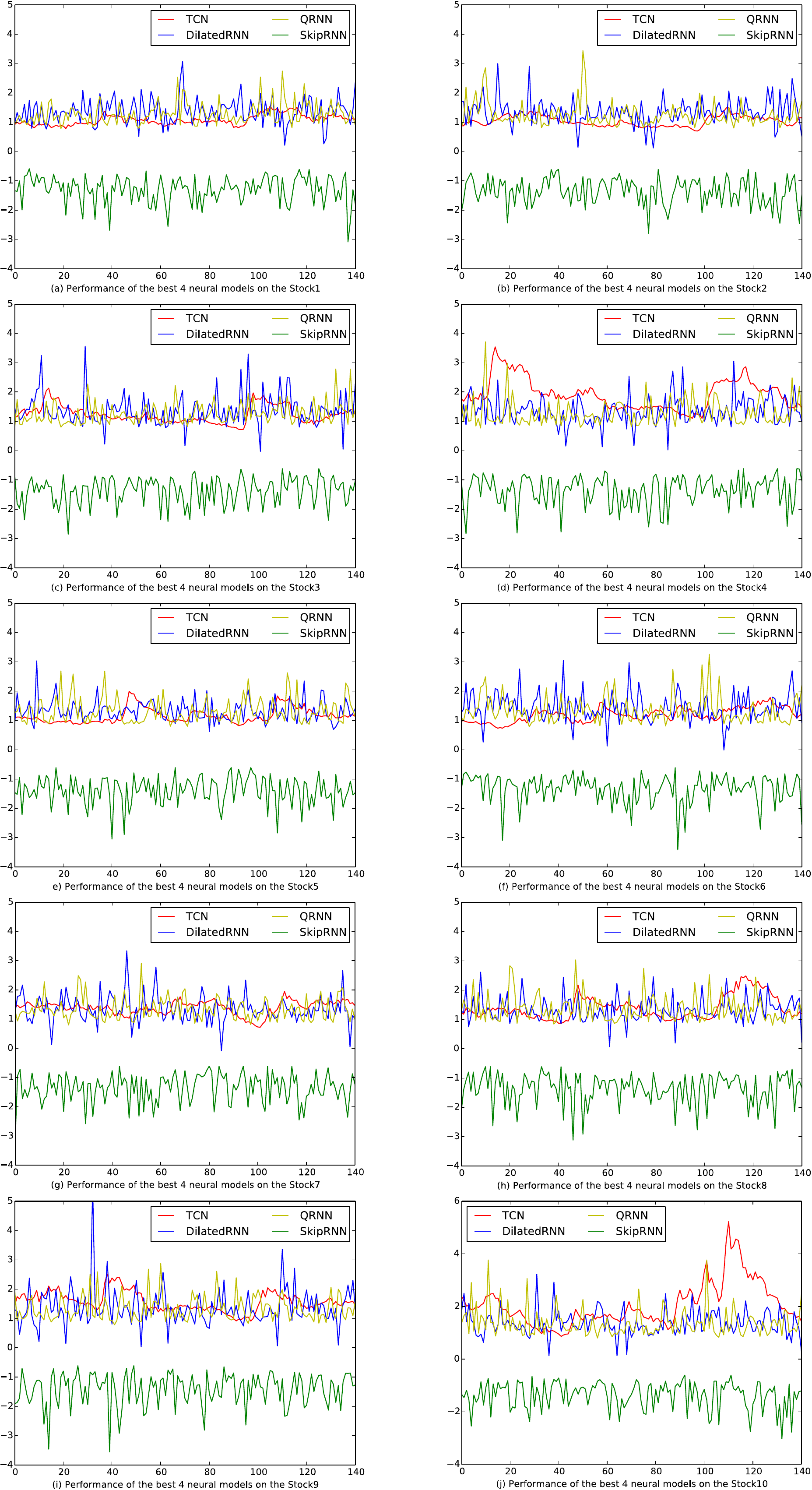}
\caption{Case study of Volatility Forecasting}
\label{fig:merged_figure}
\end{figure}

\end{document}